\documentclass{article}
\usepackage{spconf,amsmath,graphicx,hyperref}
\usepackage{algorithmic}
\usepackage{algorithm}
\usepackage{amsmath,amssymb}

\usepackage{float} 
\usepackage{multirow}

\usepackage{array}
\usepackage[caption=false,font=normalsize,labelfont=sf,textfont=sf]{subfig}
\usepackage{textcomp}
\usepackage{stfloats}
\usepackage{url}
\usepackage{verbatim}
\usepackage{graphicx}
\usepackage{xcolor}
\usepackage{makecell} 
\usepackage{cite}
\usepackage{booktabs}

\title{CrackForward: Context-Aware Severity-Stage Crack Synthesis for Data Augmentation}
\name{Nassim Sadallah and Mohand Sa\"id Allili }
\address{LARIVIA Lab, Department of Computer Science and Engineering\\
Gatineau, QC, Canada\\
\{sadn12, mohandsaid.allili\}@uqo.ca}

\begin{document}
%
\maketitle
\begin{center}
\small © 2026 IEEE. Personal use of this material is permitted. For any other use, permission must be obtained from IEEE.
\end{center}
\begin{abstract}
Reliable crack segmentation is vital for structural health monitoring, yet the scarcity of well-annotated data constitutes a major challenge. To address this limitation, we propose a novel context-aware generative framework designed to synthesize realistic crack growth patterns for data augmentation. Unlike existing methods that primarily manipulate textures or background content, CrackForward explicitly models crack morphology by combining directional crack elongation with learned thickening and branching.
Our framework integrates two key innovations: (i) a contextually guided crack expansion module, which uses local directional cues and adaptive random walk to simulate realistic propagation paths; and (ii) a two-stage U-Net-style generator that learns to reproduce spatially varying crack characteristics such as thickness, branching, and growth. 
Experimental results show that the generated samples preserve target-stage saturation and thickness characteristics and improve the performance of several crack segmentation architectures. These results indicate that structure-aware synthetic crack generation can provide more informative training data than conventional augmentation alone.

\end{abstract}
\begin{keywords}
Crack Expansion, Local Eigenvector Orientation, Unet-style generation, Crack Segmentation
\end{keywords}
\section{Introduction}
\label{sec:intro}
Rapid advances in machine learning and robotics are reshaping structural health monitoring (SHM) by enabling automated inspection, progress tracking, and proactive safety assessment of critical infrastructure \cite{10452281,halder2023robots,prunella2023deep}. Maintenance alone represents 50–70 \% of a project’s lifecycle cost \cite{sulaiman2021utilizing}, making early damage detection crucial for cost-effective asset management. Among deterioration mechanisms, concrete crack propagation is the most prevalent and damaging \cite{zeng2024meso}. Although cracks are visually observable, their early signs are often subtle and easily missed, delaying timely intervention.

To quantify deterioration, crack segmentation, which classifies each pixel as crack or background, offers precise geometric information for assessing severity, modeling growth, and planning repairs \cite{liu2019deepcrack,panella2022semantic}. However, the effectiveness of deep learning–based segmentation is constrained by limited annotated datasets, a bottleneck that hampers model generalization across diverse materials, lighting conditions, and crack morphologies \cite{rezaie2020comparison}. Addressing this data scarcity is therefore critical for building robust, scalable SHM solutions that can reliably detect and track structural damage under real-world conditions.

To mitigate data scarcity, research has explored two main dataset-expansion strategies. The first uses traditional augmentation, geometric transformations \cite{zhou2023deep}, mixup \cite{tian2021new}, and image-processing variations, to boost model robustness and generalization \cite{yang2022image,zhang2020does}. While helpful, these methods merely reshape existing samples, offering limited novelty and diversity \cite{zhang2023expanding}. A second direction leverages generative models, which synthesize images from segmentation masks or learned feature representations \cite{jin2023establishment,fontanini2025semantic}. 
Although more advanced, current approaches face two main challenges: (1) limited informativeness, as generated data often lacks contextual richness and structural diversity, and (2) a synthetic–real domain gap that causes segmentation models trained only on synthetic images to underperform unless combined with real data. In addition, most methods treat cracks as purely visual artifacts, overlooking material-specific propagation. For example, crack growth in pavement differs significantly from that in walls, making contextual differentiation essential for generating data that reflects realistic crack evolution.

In this work, we present a novel two-phase framework for context-aware crack expansion under structural and severity-stage constraints. Our main contributions are: (1) a local endpoint estimation and directional propagation strategy for structure-consistent crack elongation, and (2) a dual-stage generative architecture for stage-conditioned crack thickening and branching synthesis. Unlike standard augmentation or generic GAN-based methods, the proposed framework models stage-consistent crack development, generating masks that capture plausible morphological progression across severity levels. The first phase performs direction-guided crack elongation, while the second models dynamic thickening and branching. This yields high-fidelity, domain-aware synthetic crack masks that improve the generalization of downstream segmentation models.

The rest of this paper is organized as follows: Section \ref{methodology} presents our proposed methodology. Section  \ref{experiments} presents our experiments for validation. Section \ref{conclusion} concludes the paper.

\section{METHODOLOGY}
\label{methodology}
\subsection{Local Endpoint Estimation (LEE)}
This stage  seeks to simulate the progressive increase in crack length. This starts by reliably identifying structurally significant endpoints in a source crack mask $M$.First, we compute the crack skeleton $S\in\{0,1\}^{H\times W}$, then crack endpoints are detected using $3\times3$ kernel filtering, $N=S*K$, where $*$ is discrete convolution and :
$$  K=\begin{bmatrix}1&1&1\\1&0&1\\1&1&1\end{bmatrix},
$$
where a point $(i,j)$ is an endpoint if it is on the crack and $N(i,j) = 1$. Let $\mathcal{C}$ be the set of all endpoints. To enforce a minimum euclidean separation \(d_{\min}\) between crack propagation positions, we iteratively select 
endpoints from $\mathcal{C}$. Let \(\mathcal{E}\) be the set of final retained endpoints, which is initially empty. For each iteration, we select the next point $\textbf{e} \in \mathcal{C} $ such that:  
\begin{equation}
\forall e'\in\mathcal{E}:\|\textbf{e}-\textbf{e} '\|_2>d_{\min}.\notag
\end{equation}

\begin{figure}[h]
    \centering
        \includegraphics[width=9cm]{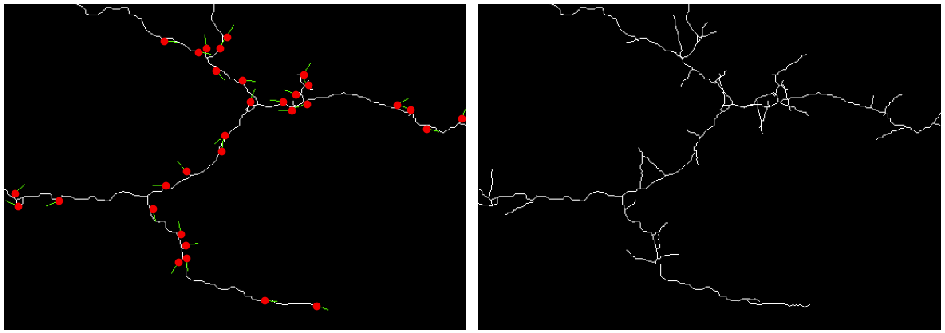}  
    \caption{Illustration of endpoint detection (left) and crack propagation using diretional random walks (right).}
    \label{fig:LEO}
\end{figure}

To estimate a crack's local orientation at endpoint candidate $\textbf{e}=(i,j)$, we first extract a centered \(w\times w\) window $M$ around the endpoint. Let $\{P_1,..., P_n\}$ be the set containing the crack pixels coordinates in \(M\),  $\mu$ the coordinates average. First, we compute the empirical covariance to capture the crack pixel spatial distribution
\[
C = \frac{1}{n-1} \sum_{k=1}^n (P_k- \mathbf{\mu})   (P_k-\mathbf{\mu})^T,
\]
and let \(\lambda_1\ge\lambda_2\) and \(\mathbf{v}_1,\mathbf{v}_2\) be its eigenpairs. The principal axis is \(\mathbf{v}_{\mathrm{dom}}=\mathbf{v}_1\). Finally, we normalize and set the direction of the crack propagation by correcting the sign of the dominant eigenvector, as follows:  
\[
\mathbf{v}_{\mathrm{adj}} =
\begin{cases}
-\mathbf{v}_{\mathrm{dom}}, & \text{if } \mathbf{v}_{\mathrm{dom}}^\top (\mu-\textbf{e}) < 0,\\[4pt]
\phantom{-}\mathbf{v}_{\mathrm{dom}}, & \text{otherwise},
\end{cases}
\quad
\mathbf{v}_{\mathrm{adj}} \leftarrow \dfrac{\mathbf{v}_{\mathrm{adj}}}{\|\mathbf{v}_{\mathrm{adj}}\|}.
\]
The estimated orientation angle is thus: $$\theta = \operatorname{atan2}\big((\mathbf{v}_{\mathrm{adj}})_y,\;(\mathbf{v}_{\mathrm{adj}})_x\big)$$
Algorithm~\ref{alg} gives the full procedure and Fig.~\ref{fig:LEO} illustrates endpoints (red) and local directions (green).

\subsection{Directional Crack Propagation}
We model crack growth from endpoints with a directional random walk balancing local orientation and controlled randomness. Let $I\in\{0,1\}^{H\times W}$ be the skeleton image and $\mathcal{E}=\{(y_i,x_i)\}$ the filtered endpoints. For each endpoint we estimate the dominant orientation $\theta_0$. Propagation is constrained to $\theta \sim \mathcal{U}[\theta_0-\delta,\theta_0+\delta]$ with step length $\ell$. Starting at $p_0$, each step updates the position by:

\begin{equation}
    p_{t+1}=p_t+\ell\,[\cos\theta,\;\sin\theta]^\top,\notag
\end{equation}
and the segment \([p_t,p_{t+1}]\) is rasterized to the image (digital line). The walk runs for \(s\) steps until a target crack density \(m\in[0,1]\) is reached. This procedure preserves local directionality while introducing controlled stochastic deviations and global sparsity control.

\begin{algorithm}[ht!]
\caption{Directional random walk algorithm}
\label{alg}
\begin{algorithmic}[1]
\REQUIRE Binary image \( I \), endpoints \( \mathcal{E} \), step range \([s_{\min}, s_{\max}]\), base length \( \ell \), max deviation \( \delta \), density \( m \)
\STATE $I_0 \leftarrow$ Skeletonize $(I)$ 
\FORALL{endpoint \( (y, x) \in \mathcal{E} \)}
    \STATE \( \theta_0, \mathbf{v}_{\text{dom}} \gets \text{LEE}(I_0, (y,x)) \)
    \STATE \( s \gets \text{random integer in } [s_{\min}, s_{\max}] \)
    \WHILE{\( s > 0 \) and \( \text{density}(I_0) < m \)}
        \STATE  \( \Delta \theta \gets \text{random uniform in } [-\delta, \delta] \)
        \STATE \( \theta_{\text{sampled}} \gets \theta_0 + \Delta \theta \)
        \STATE \( \ell_{\text{step}} \gets \ell \)
        \STATE \( x' \gets x + \ell_{\text{step}} \cdot \cos(\theta_{\text{sampled}}) \)
        \STATE \( y' \gets y + \ell_{\text{step}} \cdot \sin(\theta_{\text{sampled}}) \)
        \IF{ \( (x', y') \notin [0,W)\times[0,H) \)}
            \STATE \textbf{break}
        \ENDIF
        \STATE Draw line from \((x,y)\) to \((x',y')\) and set pixels to 1.
        \STATE \( x \gets x',\; y \gets y' \), \; \( s \gets s - 1 \)
    \ENDWHILE
\ENDFOR
\RETURN \( I_0 \)
\end{algorithmic}
\end{algorithm}

\subsection{Dual-Stage Framework for Crack Growth Synthesis}
In low-reinforced concrete, crack growth is nonlinear: as cracks propagate, their width generally increases due to ongoing material degradation and stress redistribution. Capturing this coupled evolution of length and thickness is essential for generating training data that reflects real structural behavior. 

\begin{figure}
\centering
\begin{tabular}{|c|}
\hline
\includegraphics[width=0.99\linewidth]{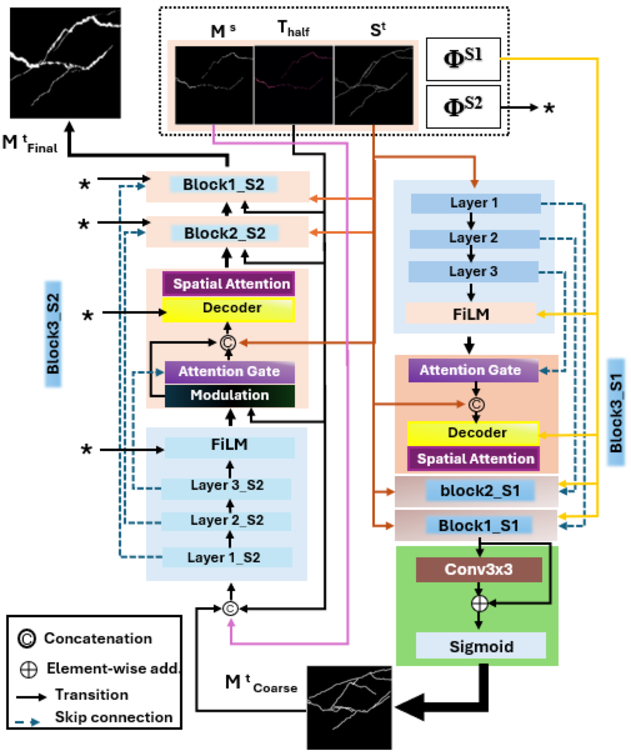}\\
\hline
\end{tabular}
\caption{Dual-stage crack thickener architecture.}
\label{fig:tanslator}
\end{figure}

To address this, we propose a dual-stage U-Net–based generator for unpaired crack-mask translation that integrates gated and dilated convolutions to synthesize realistic crack geometry with spatially varying thickness. Fig. \ref{fig:tanslator} presents the global architecture of this model.

\begin{figure}
\centering
\begin{tabular}{|c|}
\hline
\includegraphics[width=0.6\linewidth]{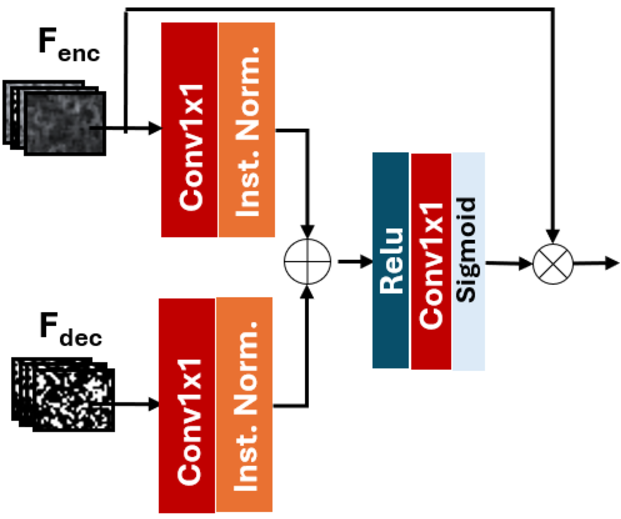}\\
\hline
\end{tabular}
\caption{Attention gate module}
\label{fig:gates}
\end{figure}

\textbf{Stage~1} models the coarse propagation path and overall length. It takes elongated skeleton \(s_{t}\in\{0,1\}^{1\times H\times W}\) and source statistics $\phi^{s1}$ and produces a coarse binary mask \(M^{t}_{coarse}\in\{0,1\}^{1\times256\times256}\). Through 3 decoding stages, FiLM layers ensure minimum thicknening,  while attention gates Fig.~\ref{fig:gates} weight skip connections to suppress irrelevant encoder informations. The decoder module uses dilated convolutions hallucinating pixels along $s_{t}$, followed by spatial attention to capture global feature relations and refine outputs.

\textbf{Stage~2} refines local widening and branching, enabling masks whose thickness naturally grows with crack extension. The generation is conditioned on the coarse mask ($M^{t}_{coarse}$), the original mask ($M^{s}$), the half-thickness map \(T^{s}_{\mathrm{half}}\), and target statistics \(\phi^{\mathrm{S2}}\).

\begin{figure}
\centering
\begin{tabular}{|c|}
\hline
\includegraphics[width=0.7\linewidth]{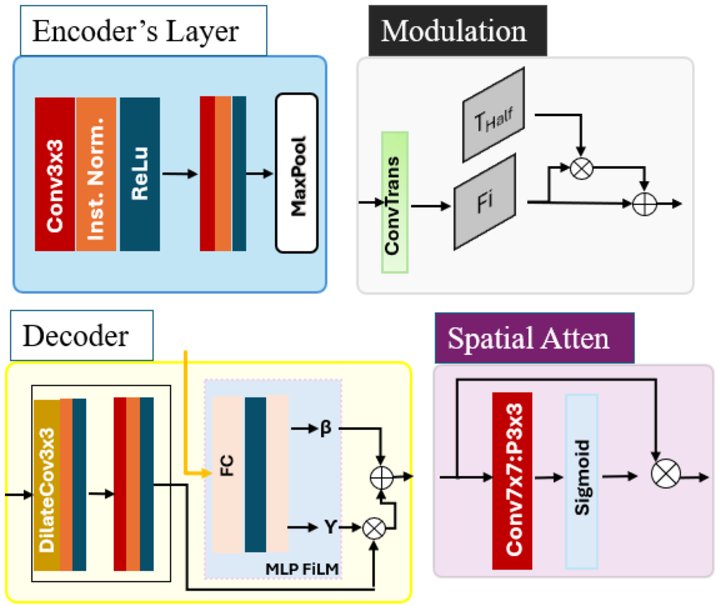}\\
\hline
\end{tabular}
\caption{modules architecture}
\label{fig:modules}
\end{figure}

The encoder comprises three convolutional blocks; an early FiLM layer \cite{perez2018film} modulates feature-wise scale and bias \((\gamma,\beta)\) using statistics \(\phi^{\mathrm{S2}}\). In the decoder, feature maps are modulated by \(T^{s}_{\mathrm{half}}\) to control per-pixel thickness adjustments. Attention gates weight skip connections to keep only relevant encoder informations, and dilated convolutions enlarge the receptive field with parameter efficiency. Decoding blocks include FiLM layers for multi-scale conditioning and a final spatial-attention module to capture global feature relations. The modulation block, spatial-attention block, encoder and decoder illustrated in Fig.~\ref{fig:modules} facilitate precise morphological control, aligning synthetic patterns with crack evolution dynamics.

\subsection{Objective functions}

We adopt an unpaired adversarial learning strategy to model crack severity progression. The generator is trained using an adversarial loss to produce realistic crack masks, together with a reconstruction loss to preserve structural consistency during growth. In addition, we introduce three morphology-aware losses, namely thickness, saturation, continuity, in order to explicitly control the geometric properties of the generated cracks and enforce realistic morphological evolution across severity stages.

We use the Least Squares GAN (LSGAN) formulation to stabilize adversarial training and improve gradient quality. Given real sample $x_i \sim p_{i}$ at stage $i$ and translated sample $\tilde{x_j} \sim \tilde{p}_{z}$ at stage $j$ with ($i, j \in \{1, \dots, N\}$ and $i<j$), the adversarial loss for the discriminator $D$ and $G$ are defined as:
\begin{eqnarray}
\label{eq:lsganD}
&\mathcal{L}_{\text{adv}}^D 
= \mathbb{E}_{x_i \sim P_{i}} \big[ (D(x_i) - 1)^2 \big] 
+ \mathbb{E}_{\tilde{x_j} \sim \tilde{P}_z} \big[ (D(\tilde{x_{j}}))^2 \big]\notag
\end{eqnarray}
\begin{align}
\label{eq:lsganG}
&\mathcal{L}_{\text{adv}}^G
&= \mathbb{E}_{x_i \sim P_{i}} \Big[ \big( D(G(x_i, T_i^{half}, s_i^{,}, \Phi_{j})) - 1 \big)^2 \Big]\notag
\end{align}

$T_i^{\mathrm{half}}$ denotes the half-thickness map of the input crack mask $M_i$, where each foreground pixel $p$ is assigned its Euclidean distance to the nearest boundary pixel, given by $T_i^{\mathrm{half}}(p)=\min_{q\in \partial M_i}\|p-q\|_2$, with $\partial M_i$ the boundary of $M_i$.
The target domain statistics $\Phi_j=[\mu_s^{(j)},\mu_t^{(j)}]$ contains the mean saturation and mean half-thickness of target stage $j$, $s_i^{,}$ is the extended skeleton generated by LEE module. The full LSGAN loss, thus, is given by $\mathcal{L}_{lsgan} = \mathcal{L}_{\text{adv}}^D + \mathcal{L}_{\text{adv}}^G $

We use the reconstruction (cycle) loss to preserve the geometric identity as we do not have paired data. The  cycle loss is defined as  $\mathcal{L}_{\text{cyc}} = \|\hat{x_i}^{\text{rec}} - x_i\|_1$.

While the adversarial loss enforces global realism, it does not explicitly constrain the morphological properties of the generated crack masks. Thus, we introduce three \emph{morphology-aware losses} that encode domain-specific structural priors: \emph{thickness}, \emph{saturation}, and \emph{continuity}.

\textbf{Thickness loss:} Penalizes deviations from the dataset mean thickness $\mu_t$: \[ \mathcal{L}_{\text{th}} = \big| \bar{t}(\hat{x_j}) - \mu_j \big|, ~~~~~~~~~~~~~ 
\bar{t}(\hat{x_j}) = \frac{1}{|\hat{x_j}|} \sum_{p \in \hat{x_j}} d(p).\]

\textbf{Saturation loss:} Enforces alignment with the target mean saturation $\mu_s$: \[
\mathcal{L}_{\text{sat}} = \big| s(\hat{x}) - \mu_s \big|, ~~~~~~~~~
s(\hat{x_j}) = \frac{1}{HW} \sum_{p} \mathbf{1}\{\hat{x_j}(p) > 0.5\},
\]

\textbf{Continuity loss:} Computed as the mean absolute Laplacian response to encourage spatial coherence and penalize fragmented structures:
\[
\mathcal{L}_{\text{cont}} = \frac{1}{HW} \sum_{p} \big| (K * \hat{x_j})(p) \big|,~~~~
K = 
\begin{bmatrix}
0 & 1 & 0 \\
1 & -4 & 1 \\
0 & 1 & 0
\end{bmatrix}.
\]

The complete generator loss combines the adversarial objective with the morphology-aware losses:
\[
\mathcal{L}_{G} 
= \mathcal{L}_{\text{lsgan}} 
+\lambda_1 \mathcal{L}_{\text{cyc}} 
+ \lambda_2 \mathcal{L}_{\text{th}} 
+ \lambda_3 \mathcal{L}_{\text{sat}} 
+ \lambda_4 \mathcal{L}_{\text{cont}},
\]
where $\lambda_1$, $\lambda_2$, $\lambda_3$, and $\lambda_4$ are weights for cycle and morphology losses.

\section{Experimental Results}
\label{experiments}

\subsection{Experiment Setup}
To validate our framework, we conducted experiments using the DeepCrack dataset \cite{liu2019deepcrack}, consisting of 537 images. The dataset was partitioned into three balanced severity stages using the 33rd and 66th percentiles of a per-sample severity score \(\phi\), computed from crack saturation and mean half-thickness. Table~\ref{tab:crack_3stage_all_stats} summarizes the statistics of each stage.

\begin{table}[h]
\centering
\caption{Per-stage crack statistics}
\label{tab:crack_3stage_all_stats}
\setlength{\tabcolsep}{6.0pt}
\renewcommand{\arraystretch}{0.80}
\small
\begin{tabular}{c c c c c}
\toprule
Stage & Split & $N$ & $s\;(\mu \pm \sigma)$ & $t\;(\mu \pm \sigma)$ \\
\midrule

\multirow{4}{*}{0}
& FULL  & 179 & $0.0116 \pm 0.0047$ & $1.550 \pm 0.359$ \\
& Train & 125 & $0.0117 \pm 0.0048$ & $1.184 \pm 0.158$ \\
& Val   & 27  & $0.0113 \pm 0.0052$ & $1.141 \pm 0.117$ \\
& Test  & 27  & $0.0113 \pm 0.0041$ & $1.222 \pm 0.212$ \\
\midrule

\multirow{4}{*}{1}
& FULL  & 179 & $0.0282 \pm 0.0070$ & $2.563 \pm 0.455$ \\
& Train & 125 & $0.0281 \pm 0.0070$ & $1.701 \pm 0.253$ \\
& Val   & 27  & $0.0294 \pm 0.0072$ & $1.695 \pm 0.311$ \\
& Test  & 27  & $0.0269 \pm 0.0068$ & $1.809 \pm 0.342$ \\
\midrule

\multirow{4}{*}{2}
& FULL  & 179 & $0.0655 \pm 0.0359$ & $5.605 \pm 2.883$ \\
& Train & 125 & $0.0663 \pm 0.0354$ & $3.509 \pm 1.692$ \\
& Val   & 27  & $0.0684 \pm 0.0452$ & $3.427 \pm 1.843$ \\
& Test  & 27  & $0.0549 \pm 0.0200$ & $2.945 \pm 1.000$ \\
\bottomrule
\end{tabular}
\end{table}

All masks and images were resized to $256\times256$. Both the generator and the segmentation models were trained with batch size 4 using the Adam optimizer with learning rate $2\times10^{-4}$. The generator was trained in an unpaired setting for 100 epochs on an NVIDIA RTX6000 GPU, with cycle loss weight $\lambda_1=10.0$ and morphology-aware loss weights $\lambda_2=2.0$, $\lambda_3=2.0$, and $\lambda_4=4.0$. In the directional propagation stage, the parameters were set to $d_{\min}=4$, $\delta=90^\circ$, $\ell=2$, and $[s_{\min},s_{\max}]=[3,50]$, with stopping controlled by the target mean saturation $m$. For downstream evaluation, all segmentation models were trained on the same split and input resolution for 30 epochs.

\subsection{Performance Evaluation}

\subsubsection{Quantitative evaluation}
In this assessment, we compared the distribution of the generated translations with the empirical distribution of the actual stages, and we report the corresponding results in Table~\ref{tab:comparative_distribution}. The proposed generator performs translations with saturation and thickness errors of approximately 2.6\% and 4.4\% for Stage~1, and 6.7\% and 0.14\% for Stage~2, respectively. These low relative deviations indicate that the growth accurately preserves the target-stage statistical properties, and remains tightly aligned with the real data distributions in terms of both crack density and structural thickness.

Table \ref{tab:ssim} reports quantitative assessments via L1 loss, SSIM, and PSNR on the generated images, revealing low reconstruction errors and high structural fidelity.

\begin{table}[!]
\centering
\caption{Per-stage real vs. generated statistics on the TEST }
\label{tab:comparative_distribution}
\setlength{\tabcolsep}{3.0pt}
\renewcommand{\arraystretch}{0.80}
\small
\begin{tabular}{c c c c c c}
\toprule
Stage & Case &
$s\;(\mu \pm \sigma)$ & $\Delta s$ &
$t\;(\mu \pm \sigma)$ & $\Delta t$ \\
\midrule
1 & Real &
$0.0269 \pm 0.0068$ &  &
$1.809 \pm 0.342$ &  \\

1 & Fake &
$0.0262 \pm 0.0034$ & $\mathbf{0.0007}$ &
$1.730 \pm 0.180$ & $\mathbf{0.079}$ \\
\midrule
2 & Real &
$0.0549 \pm 0.0200$ &  &
$2.945 \pm 1.000$ &  \\

2 & Fake &
$0.0512 \pm 0.0119$ & $\mathbf{0.0037}$ &
$2.941 \pm 0.881$ & $\mathbf{0.004}$ \\
\bottomrule
\end{tabular}
\end{table}

\begin{table}[!]
\centering
\caption{Quantitative assessment of synthesized cracks.}
\label{tab:ssim}
\small
\setlength{\tabcolsep}{6pt}
\begin{tabular}{l@{\hskip 4pt}ccc}
\hline
\textbf{domain target} & \textbf{L1 Loss} & \textbf{SSIM} & \textbf{PSNR (dB)} \\
\hline
\textbf{Medium cracks (stage 1)}                   & 0.1001 & 0.968 & 32.62 \\
\textbf{Advanced cracks (stage 2)}            & 0.1073 & 0.9510 & 30.84 \\

\hline
\end{tabular}
\end{table}

\subsubsection{Qualitative evaluation}
\begin{figure*}[htbp!]
\centering
\includegraphics[width=0.99\linewidth]{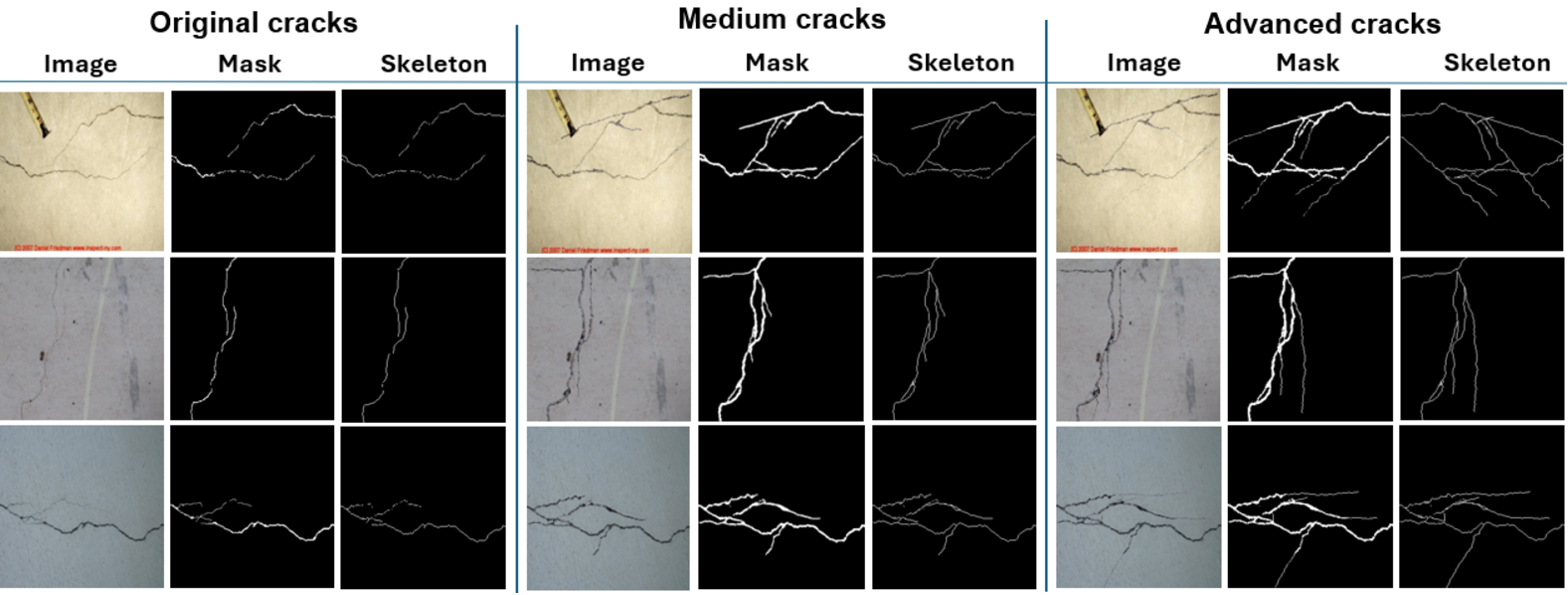}
\caption{Representative samples of crack propagation synthesized by the proposed CrackForward framework.}
\label{fig:growthExample}
\end{figure*}

Figure~\ref{fig:growthExample} shows representative examples of crack propagation synthesized by CrackForward, illustrating plausible elongation patterns and spatially varying thickness. In particular, the generated samples exhibit stronger thickening near existing branches and thinner newly elongated regions, in agreement with the intended stage-wise morphological progression. These qualitative results further support the usefulness of the proposed framework for crack data augmentation.

\subsubsection{Segmentation application enhancement}
To evaluate CrackForward for crack segmentation, U\mbox{-}Net, FPN, PAN, and PSPNet were trained under three settings: real data only, real data with classical augmentation, and real data with CrackForward augmentation. In all cases, the real training set included the three severity stages. CrackForward added synthetic samples by translating early-stage masks to medium and severe stages, and medium-stage masks to the severe stage. Classical augmentation used random flips, rotations, zooms, color jittering, brightness, and hue variations.
As shown in Table~\ref{tab:segmentation_results}, CrackForward consistently improves performance across all models. Relative to real-only training, Dice improves by 4.7\%, 3.7\%, 6.6\%, and 3.4\% for U\mbox{-}Net, FPN, PAN, and PSPNet, respectively, while IoU improves by 6.3\%, 5.4\%, 7.9\%, and 5.8\%. Compared with classical augmentation, Dice further improves by 3.3\%, 2.9\%, 4.2\%, and 2.6\%, and IoU by 5.4\%, 4.9\%, 5.7\%, and 4.6\%, respectively. These results indicate that CrackForward provides complementary structural variability beyond conventional transformations.

\begin{table}[h!]
\centering
\caption{Segmentation performance under different augmentation settings.}
\label{tab:segmentation_results}
\setlength{\tabcolsep}{9pt}
\renewcommand{\arraystretch}{0.7}
\small
\begin{tabular}{c c c c}
\toprule
Model & Training Data & Dice & IoU \\
\midrule

\multirow{3}{*}{U-Net} 
& Real & 0.771 & 0.663 \\
& Real + Classical Aug. & 0.781 & 0.669 \\
& Real + CrackForward &  \textbf{0.807} & \textbf{0.705}  \\
\midrule

\multirow{3}{*}{FPN} 
& Real & 0.725 & 0.589 \\
& Real + Classical Aug. & 0.731 & 0.592\\
& Real + CrackForward & \textbf{0.752} & \textbf{0.621} \\
\midrule

\multirow{3}{*}{PAN} 
& Real & 0.701 & 0.567 \\
& Real + Classical Aug. & 0.717 & 0.579 \\
& Real + CrackForward & \textbf{0.747} & \textbf{0.612} \\
\midrule

\multirow{3}{*}{PSPNet} 
& Real & 0.715 & 0.582 \\
& Real + Classical Aug. & 0.720 & 0.589 \\
& Real + CrackForward & \textbf{0.739} & \textbf{0.616}  \\
\bottomrule
\end{tabular}
\end{table}

\section{CONCLUSIONS}
\label{conclusion}
This paper introduced CrackForward, a context-aware generative framework for alleviating data scarcity in crack segmentation. By combining context-guided crack expansion with a two-stage U-Net-style generator, the proposed method synthesizes stage-consistent crack patterns with controlled thickness, saturation, and structural variability. Experimental results show that the generated samples preserve target-stage morphological characteristics and consistently improve segmentation performance across multiple architectures.

Overall, the results suggest that structure-aware crack synthesis can provide more informative augmentation than conventional transformations alone. Future work will investigate broader experimental validation, joint modeling of crack masks and background appearance, and extension to other structural defects such as spalling, delamination, and corrosion.

\section*{Acknowledgments}
This research was supported by the Fonds de recherche du Québec -- Nature et technologies (FRQNT), award DOI: 10.69777/371547.

\bibliographystyle{IEEEbib}
\bibliography{references}

\end{document}